\pdfoutput=1

\documentclass[11pt]{article}

\usepackage[]{ACL2023}

\usepackage{times}
\usepackage{latexsym}

\usepackage[T1]{fontenc}

\usepackage[utf8]{inputenc}

\usepackage{microtype}

\usepackage{inconsolata}

\usepackage{booktabs}
\usepackage{tikz}
\usepackage{pgfplots}
\usepackage{bbm}
\pgfplotsset{compat=newest}
\usepgfplotslibrary{colorbrewer}

\title{Style Locality for Controllable Generation with kNN Language Models}

\author{Gilles Nawezi $\dagger$ \and Lucie Flek $\dagger \ddagger$ \and Charles Welch $\ddagger$\\
    Conversational AI and Social Analytics (CAISA) Lab \\ 
    $\dagger$ Department of Mathematics and Computer Science, University of Marburg \\ 
    $\ddagger$ Bonn-Aachen International Center for Information Technology (b-it), University of Bonn \\
}

\begin{document}
\maketitle
\begin{abstract}
Recent language models have been improved by the addition of external memory. Nearest neighbor language models retrieve similar contexts to assist in word prediction. The addition of locality levels allows a model to learn how to weight neighbors based on their relative location to the current text in source documents, and have been shown to further improve model performance. Nearest neighbor models have been explored for controllable generation but have not examined the use of locality levels. We present a novel approach for this purpose and evaluate it using automatic and human evaluation on politeness, formality, supportiveness, and toxicity textual data. We find that our model is successfully able to control style and provides a better fluency-style trade-off than previous work.
\end{abstract}

\section{Introduction}


Controllable language generation is the study of developing models to generate language with predefined characteristics, including stylistic (e.g. politeness, formality), demographic (e.g. gender, age), or content-related (e.g. keywords, entities) attributes~\cite{Prabhumoye.04.05.2020}.
In recent years, chatbots have quickly gained widespread use in industries including marketing, support systems, education, healthcare, cultural heritage, and entertainment~\cite{Adamopoulou.2020}. Chatbots generate language to prepare a human-like or natural response for the user based on the intent and context information provided by the user, environment, and other factors \cite{S.Singh.2016}. For a chatbot in a support system, controllable generation could be leveraged to respond more politely. In an educational context, the chatbot could give more entertaining factual answers to enhance the learning experience. 


Recently, memory-augmented language models have been shown to substantially improve performance by memorizing rare language patterns~\cite{Khandelwal.01.11.2019,trotta-etal-2022-nearest}. The use of a datastore allows the model to scale to larger text collections without additional training by adding new data to the datastore.
We present a model that combines the advantages of transformer language models, locality types, and memorization to enable the controlled generation of language in particular styles. 
The model makes use of a k-nearest neighbor architecture to retrieve similar contexts from memory and uses specifically trained locality weights to reweight neighbors according to the desired style. In contrast to previous locality work, our locality levels are based on the style, data source, and similarity between styles, rather than the location of text within a document~\cite{Xu.06.10.2021}. We compare our model to previous work, showing that it outperforms previous memory-augmented language models on style control, while providing a favorable trade-off between style control and fluency.

\section{Related Work}

\textbf{Language Models with Memorization}
\newline
Traditional neural network language models have achieved state-of-the-art performance in the field of language modeling, but they are unable to change in response to recent events. As a result, they are unable to adapt to changing environments \cite{Dodge.21.11.2015}. 
The addition of an external memory to these networks is a method for resolving this issue. These models are able to store data in their external memory and adjust to a changing patterns \cite{Grave.14.12.2016}.

Neural Turing Machines are neural networks connected to resources in external memory that they can access through attention. The combined system is comparable to a Turing machine but is differentiable and can be trained by gradient descent. Work from \citet{Graves.20.10.2014} showed the model's ability to learn algorithms like copying, sorting, and associative recall.
Memory-based recurrent neural networks similarly have been shown to implement continuously differentiable versions of well-known data structures including stacks and queues, outperforming deep RNNs on sequence-to-sequence transduction tasks \cite{Grefenstette.08.06.2015}, and can solve difficult pattern recognition tasks with an expandable memory~\cite{Joulin.03.03.2015}.



End-to-end memory networks are neural networks that have a recurrent attention model applied to a potentially huge external memory. Since the model is trained end-to-end, it needs much less supervision during training and is therefore more widely applicable in realistic settings. On the Text8 and Penn Treebank \cite{MitchellPMarcus.2002} datasets, the model showed similar performance to RNNs and LSTMs \cite{Sukhbaatar.31.03.2015}. 

Extending language models with a continuous cache to adjust their prediction to the recent history showed improvement over previous memory-augmented networks~\cite{Grave.14.12.2016}. The model is a more condensed version, which accesses past hidden activations through a dot product with the current hidden activation and stores them as memory. This system is effective in handling enormous memory sizes.

\citet{Khandelwal.01.11.2019} introduced k-nearest neighbor language models (kNN-LMs), outperforming previous memory approaches. The model encodes context vectors and stores them as keys in a dictionary with a corresponding next word. During decoding, the model finds nearest neighbors to the current context, retrieves their values from the dictionary, and interpolates the distribution given by the nearest neighbors with the distribution given by the base language model.

Recently, \citet{Xu.06.10.2021} introduced structural locality levels to kNN-LMs. The idea is to reweight nearest neighbors by their similarity to other text as determined by the location of the text in the structure of the document. They showed that their model outperformed \citet{Khandelwal.01.11.2019} on Wikipedia text and Java code. The locality levels for code included the project and directory that the code was in, while for Wikipedia it included four levels; (1) different category and section, (2) same section - different category, (3) different section - same category, and (4) same category and section. Each of these levels is assigned a weight, learned by a linear model on a sample of the data. They found that learned weights are lower for more similar levels and that the final predictions are more likely to be nearest neighbors from the same locality.
\newline
\newline
\textbf{Controllable Text Generation}
\newline
Controllable characteristics of text include stylistic ones like formality, sentiment, and politeness. They may also be the demographic characteristics of the author, such as age and gender, or keywords and entities as content \cite{Prabhumoye.04.05.2020}.

\citet{Ziegler.18.09.2019} showed that language models can control use of sentiment and physically descriptive language when learning from human feedback via reinforcement learning. \citet{Dathathri.04.12.2019} introduced a model where attribute classifiers are used to guide generation. The gradients of these classifiers are used to push the latent space representation to control for the topical style, sentiment, and toxicity. \citet{Zellers.29.05.2019} created a generator and detector of fake news that controls generated text by conditioning on the authors, date, headline, and/or domain.




Other work examines control at decoding time. \citet{Liu.07.05.2021} used a combination of LMs fine-tuned on desirable or undesirable properties. The distribution of next word probabilities from the desired properties are added to the base LM, while the undesirable distribution is subtracted. Similarly, \citet{Yang.} used classifiers to determine if possible generation steps will result in text with a given stylistic attribute.



\citet{trotta-etal-2022-nearest} used a kNN language model which was modified to encode style attributes, altering the encoding space such that retrieved neighbors are more often posts of a particular style. They experimented with politeness, toxicity, and formality, finding that their architecture provides better style control and fluency for some styles over others, and that specialized datastores for each style outperformed datastores with multiple styles.

\section{Datasets}

The data collected for the experiments were obtained from four different style categories; supportive, formal/informal, polite/impolite and toxic data. In subsequent experiments, we use all combined style datasets excluding neutral data except for experiments on single style models. All data is in English.

\noindent
\textbf{Supportive}
\citet{Hada.10.06.2021} published the \textit{Ruddit} corpus, containing 6k Reddit comments and a continuous value representing the offensiveness or supportiveness of each in the range of -1 to 1. We took the top 25\% of the dataset as the supportive data, representing 1.1k comments.




\noindent
\textbf{Formal / Informal}
\citet{Rao.17.03.2018} introduced the GYAFC corpus containing $\approx$ 135k informal and $\approx$ 144k formal sentences for style transfer. The documents cover the topics \textit{Entertainment and Music} and \textit{Family and Relationships}.


\noindent
\textbf{Polite / Impolite}
The politeness corpus comes from \citet{DanescuNiculescuMizil.25.06.2013} and contains 4k posts from Wikipedia Talk pages. Each comment has a label for polite, impolite, or neutral, with 1k of each polite and impolite.


\noindent
\textbf{Toxic}
We use two datasets for toxicity. The Wikipedia Detox corpus~\cite{NithumThain.2017} contains 116k comments that were labeled for toxicity from Wikipedia talk pages, with 14k labeled as toxic.
The second dataset comes from the Civil Comments Platform~\cite{Jigsaw.2019} and contains 160k toxic comments. The dataset contains toxicity subcategories that were not used in this work.

\section{Models}

Our work builds off of the kNN language models used in previous work. In a standard language model, we want to predict the next word $w_t$ given a context $w_1,...,w_{t-1}$. The language model learns the probability $p(w_t|w_1,...,w_{t-1})$.

A kNN language model requires a datastore, $D$, which is constructed with the encodings of context sequences and their corresponding target word. The encodings are given by a function, $f$, which in our work is computed with a transformer language model. The keys, $K$, and values, $V$ of the datastore are then defined as $(K,V)=\{(f(w_1,...,w_{t-1}),w_t)\}$ for all context and target word pairs in $D$.

During inference, the model must retrieve similar contexts to the current context. The current context sequence is encoded with $f$ and used to query the datastore to retrieve the nearest neighbors using a distance function, which is the $L^2$ distance in our experiments. These distances are converted into a distribution by passing the distances to a softmax and aggregating across duplicate target words. Finally, the model interpolates the nearest neighbor distribution with the base language model distribution with hyperparameter $\lambda$, resulting in the final distribution $p(w_t|w_1,...,w_{t-1})=\lambda p_{kNN} + (1-\lambda) p_{LM}$.

\citet{Xu.06.10.2021} introduced locality levels on top of this model, intuiting that additional information about the documents that the text came from could be used to weight nearest neighbors differently. The information they use for this can be thought of as a kind of depth in structured documents. For example, text from the same article and section on Wikipedia should probably be more similar than text from different articles or even different sections. We also use locality levels, but do not associate these levels with depth but rather with specific and general styles, and with the source of the text (e.g. Wikipedia versus Yahoo answers).

 The kNN distribution can be defined formally as $p_{kNN}(w_t|c_t) \propto \sum_{(k_i,v_i)\in N} \mathbbm{1}_{w_t=v_i} exp(-d(k_i,f(c_t)))$ for key-value pairs $(k,v)$ in the set of nearest neighbors, $N$, distance function, $d$, and context $c_t=w_1,...,w_{t-1}$. The addition of locality levels is a parameterization of the distance function, which reweights the distances based on the similarity of localities between the neighbor contexts and current context. This changes the exponentiated function and distribution to $p_{kNN}(w_t|c_t;\theta_n) \propto \sum_{(k_i,v_i)\in N} \mathbbm{1}_{w_t=v_i} exp(-g(k_i,f(c_t);\theta_n))$ for the function $g(k_i,c_t;\theta_n)=g_n(d(k_i,f(c_t));\theta_n)$ for key and context sharing locality level $n$. We use a linear function for $g$, as in previous work, with a bias term of zero. The function can be learned with an annotated sample of the training data and is optimized to minimize the log-likelihood of the target token.

 \citet{trotta-etal-2022-nearest} experimented with using kNN language models for controllable generation and included two baselines. The first is single style models, which are trained on only one style each. This is less efficient as it requires training a separate model for each style. The other baseline is a single style datastore approach. In this method, there are separate datastores for each style rather than having all styles mixed into one datastore. Intuitively, the model may benefit from having more similar neighbors in the datastore from which to interpolate, but these similar contexts may not share the desired style to a satisfactory degree. Using a separate datastore for each style and restricting retrieval to the desired datastore is effectively forcing the weights of $g$ to zero for contexts that do not share the same style locality level.

\section{Experimental Setup}
We first built a language model from the style dataset and compared it to a fine-tuned language model first trained on WikiText103. All models from previous work were reimplemented and fine-tuned on our style dataset. We compare to single style models, which are models fine-tuned on only the subset of data for that particular style, which means one model is fine-tuned for each style subset. Additionally, we compare to the specific datastore setting from \citet{trotta-etal-2022-nearest}. This means that instead of encoding all contexts in a single datastore we instead store them separately, so that retrieval is forced to return only neighbors in that style.
We describe our training of the style locality generation model and our comparison using human evaluation.



Experiments used an NVIDIA A100 GPU. The datastores for our experiments took between 8GB and 20GB disk space.
Aside from the datastore sizes, we use the same parameters as \citet{Khandelwal.01.11.2019}.
We built our models off of the FairSeq library \cite{Ott.01.04.2019}. We used their wiki103 LM architecture.\footnote{\url{https://github.com/facebookresearch/fairseq/blob/main/fairseq/config/model/transformer_lm}} The model has a hidden size of 1024, 16 layers, and 8 attention heads.


\subsection{Language Modeling}
We trained the \textit{wiki103} language model from Fairseq on the style data for 200 epochs with early stopping and achieved a perplexity of 134.
We then first trained the model on WikiText-103 \cite{Merity.26.09.2016} before fine-tuning on the style data, resulting in a perplexity of 62. When doing so, around 20\% of tokens in the style dataset were not in the vocab. We adjusted the vocab to include new tokens from the style datasets and to prevent the vocab from greatly expanding we changed the min count from 1 to 5, resulting in a vocabulary of 208k tokens, 60k fewer than the original. The perplexity of the model increases slightly to 68, as it now has more known tokens to predict (1.5\% unknown). Next, we implemented the k-nearest neighbors approach of \citet{Khandelwal.01.11.2019} using the same fine-tuned language model. This resulted in a perplexity six points lower, at 62, as shown in the first row of Table~\ref{Tab:perplexity_locality}. This model served as the basis of the style locality models discussed in the following section.

\begin{table}[]
    \centering
    \small
    \begin{tabular}{lc}
     \toprule
     \textbf{kNN with Locality Features} & \textbf{Perplexity}\\
     \midrule
        None \cite{Khandelwal.01.11.2019} & 62.20 \\
        Style Only & 61.89 \\
        Category Only & 62.07 \\
        Source Only & 61.86 \\
        Style \& Category & 61.91 \\
        Source \& Category & 66.76 \\
        Style \& Source & \textbf{61.73} \\
     \bottomrule
    \end{tabular}
    \caption{
    Fine-tuned transformer models with kNN component and structural locality features. Bold indicates the highest performing model.
    }
    \label{Tab:perplexity_locality}
\end{table}

\begin{table*}[]
    \centering
    \small
    \begin{tabular}{l ccc ccc c}
    \toprule
     & \multicolumn{3}{c}{Style Control} & \multicolumn{3}{c}{Fluency} & \\
    \textbf{Style} & \textbf{Model} & \textbf{Equal} & \textbf{Style Locality} & \textbf{Model} & \textbf{Equal} & \textbf{Style Locality} & \textbf{Combined} \\
    \midrule
    \citet{Khandelwal.01.11.2019} & 24\% & 31\% & \textbf{45\%} & \textbf{40\%} & 38\% & 21\% & +2\%  \\
    Single Style Model  & 27\%  & 35\%  & \textbf{37\%} & 24\%  & 34\%  & \textbf{42\%}  & +14\% \\
    Single Style Datastore  & \textbf{29\%}  &  48\% &  23\% & 16\% &  49\% &  \textbf{36\%} & +7\% \\
    \citet{trotta-etal-2022-nearest} & \textbf{39\%} &   28\% &    33\% & 8\%  &   13\%  &    \textbf{79\%}  & +33\% \\
    \bottomrule
    \end{tabular}
    \caption{Human evaluation for style control (left) and fluency (right) showing the percent of times our style locality model was chosen over the model on the left, n=50 questions per style. Combined column indicates the absolute percentage of times style locality model was chosen over the comparison model for both fluency and style control.} 
    \label{Tab:style_style_locality_vs_models}
\end{table*}

\subsection{Training the Adaptive Weights}
In order to integrate the stylistic information into the model, the distance function for determining the nearest neighbors is modified using learned weights. The locality weights are created with features representing the style, source, and category of the data. The style feature represents the text style (e.g. polite or impolite), and has nine possible values including neutral. The source represents the dataset that the data came from and includes five values, one for each dataset.
The category feature represents the style type and includes two values; positve, negative (e.g. polite and supportive are positive, impolite and toxic are negative). The categories were derived from an analysis of the similarity of subsets of our data provided in Appendix~\ref{sec:app_SBERT}.
A subset of 100k samples from the training data is used to learn the adaptive weights, i.e. the linear model for reweighting nearest neighbors based on these features. The linear model takes each neighbor's context encoding, distance, and locality features as input and returns a modified distance. The locality is encoded with a one-hot vector representing the locality combination. For example, for source and style there are four locality combinations, e.g. same source and different style, and only two possibilities when only one locality level is used.

The results in Table~\ref{Tab:perplexity_locality} show that the best performing model uses the style and source localities together. Note that the perplexities here are higher than those reported in \citet{Khandelwal.01.11.2019}, which is due to the different data used for testing. The stylistic data from social media is significantly different from the Wikipedia data that was used in previous work. Overall, this is a small improvement over the kNN model from previous work which uses no locality information. However, we performed a human evaluation, as perplexity is known to correlate poorly with human judgements~\cite{kuribayashi-etal-2021-lower,meister-cotterell-2021-language}.



\subsection{Human Evaluation}\label{sec:human_eval}

Human surveys were conducted via Limesurvey.org to compare the performance of our models. For each comparison, a total of 300 questions were asked, evenly distributed across each six styles. Each question contained the beginning of a sentence randomly selected from the dataset. Sentences were split in half and both our style locality model and the comparison model were used to complete the sentence, similar to the evaluation from \citet{gehman-etal-2020-realtoxicityprompts}.


Surveys were conducted to compare the style locality model to \citet{Khandelwal.01.11.2019}, single style models, mixed and single style datastores, and to \citet{trotta-etal-2022-nearest}. We used our best locality model for this comparison; trained with data source and style localities. Participants were shown the beginning of sentences and asked which of two continuations was more fluent and which more closely matched the desired style. The results in Table~\ref{Tab:style_style_locality_vs_models} show how often each model was preferred. We see that the single style datastore and \citet{trotta-etal-2022-nearest} show better style control, though they perform poorly on fluency. In fact, although \citet{trotta-etal-2022-nearest}'s model shows strong style control, it greatly suffers in fluency such that it may not be usable for many applications. There is always a trade-off between the two, but our \textit{combined} column shows the absolute percentage of times that our style locality model was preferred. Our model outperforms all previous work, though the closest comparison is to the kNN model of \citet{Khandelwal.01.11.2019}. We see strong style control from our model here, but at the cost of fluency. 
We find that the single style models perform worse than our model in both fluency and style and contrary to previous work, single style datastores did not show a clear advantage, with similar style control and much worse fluency.



\section{Conclusion}

We constructed a k-nearest neighbors language model that is able to leverage locality levels based on the style, data source, and similarity between styles, effectively reweighting nearest neighbors with additional available information. We discussed how our model is trained using polite, impolite, formal, informal, supportive, and toxic data.
Reimplementing models from previous work, we found that using style and data source gave an improved perplexity and was able to outperform the single style models and \citet{Khandelwal.01.11.2019} on style control. In contrast to previous work, we found that the single style datastores were only slightly better for style-control but came with a large drop in fluency, making them less favored overall. Our model showed better fluency than all models except for \citet{Khandelwal.01.11.2019}, though our model provides a better style-fluency trade-off than previous work. Our work provides initial steps toward style controllable nearest neighbors language models. Future work may explore the style-fluency trade-off in more detail and the impact on downstream applications, such as with conversational agents.

\section*{Limitations}
Our method required the use of a high-end GPU to carry out our experiments. We largely used hyperparameters chosen in the previous work we built off of due to the high compute required for this task. While our model shows a favorable style-fluency trade-off, it is possible that tuning some of the parameters from \citet{trotta-etal-2022-nearest} would lead to a more favorable trade-off. Our work fine-tunes on a relatively small set of style data which resulted in perplexities higher than results for many recent large language models. It is possible that with additional data and training time that the perplexities would be much lower, making the difference between models indistinguishable by automatic metrics.

\section*{Ethics Statement}

The development of controllable generation models, such as the one presented in this paper, has enabled the generation of text in various styles. However, it is crucial to exercise caution when using these models, as they have the potential to perpetuate harmful biases and reinforce societal power imbalances \cite{Bender.2021}. While the use of supportive or polite language is generally harmless, the irresponsible use of toxic or impolite language can cause harm. It is essential to consider the potential risks and consequences of using language models and ensure their responsible and ethical use. The work presented in this paper is intended for scientific purposes only and we recommend against the use of language models trained on harmful language data in downstream applications.


\bibliography{custom}
\bibliographystyle{acl_natbib}

\appendix

\begin{table*}[]
    \centering
    \begin{tabular}{p{13.5cm}c}
     \toprule
     \textbf{Comment} & \textbf{Style} \\
     \midrule
     So so awesome. Really want to see more. & Supportive  \\
     \midrule
    Mad dog will surely put the liberals in mental hospitals. Boorah & Toxic\\ 
    \midrule
     The In-Laws movie isn't a holiday movie, but it's okay. & Formal\\ 
     \midrule
    of corse i be wachin it evry day, my fav charachter is Inuasha & Informal  \\
    \midrule
    Ok, that's no problem. Can you recommend any other users who would be able to train me?
     & Polite\\ 
      \midrule
    I asked you a question. Be educated and please respond: What does <url> has  to do with <url>?
     & Impolite  \\
     \bottomrule
    \end{tabular}
    \caption{Samples from the Ruddit (supportive, offensive), Wikipedia Detox and Civil Comments (toxic), GYAFC (formal, informal), and Stanford Politeness (polite, impolite) datasets showing sample comments and posts.}
    \label{tab:data_samples}
\end{table*}

\section{Data Samples}
\label{sec:app_data_samples}

\textcolor{red}{\textit{Content Warning}: This paper includes examples in the appendix of language that may be offensive and upsetting.}

This section contains examples of each style for the datasets we used in our experiments. See Table~\ref{tab:data_samples} for samples from each dataset and style.

\section{Categories Derived from SBERT Analysis}\label{sec:app_SBERT}

\begin{figure}[t]
    \centering

\scalebox{.9}{%
    \begin{tikzpicture}
      \begin{axis}[enlargelimits=false,width=6.7cm,colorbar,colormap/Greys,
    xtick = {0,1,2,3,4,5},
    xticklabels = {formal,impolite,informal,polite,supportive,toxic},
    ytick = {0,1,2,3,4,5},
    yticklabels = {formal,impolite,informal,polite,supportive,toxic},
    xticklabel style={rotate=90},
    every node near coord/.append style={yshift=-0.25cm} 
      ]
        \addplot [matrix plot,point meta=explicit,]
          coordinates {
(0,0) [1] (1,0) [.38] (2,0) [.94] (3,0) [.25] (4,0) [.75] (5,0) [.43]

(0,1) [.38] (1,1) [1] (2,1) [.33] (3,1) [.84] (4,1) [.54] (5,1) [.54]

(0,2) [.94] (1,2) [.33] (2,2) [1] (3,2) [.2] (4,2) [.75] (5,2) [.39]

(0,3) [.25] (1,3) [.84] (2,3) [.2] (3,3) [1] (4,3) [.46] (5,3) [.29]

(0,4) [.75] (1,4) [.54] (2,4) [.75] (3,4) [.46] (4,4) [1] (5,4) [.55]

(0,5) [.43] (1,5) [.54] (2,5) [.39] (3,5) [.29] (4,5) [.55] (5,5) [1]
 
          };
        \node[text=black] at (axis cs: 1,0) {38};
        \node[text=white] at (axis cs: 2,0) {94};
        \node[text=black] at (axis cs: 3,0) {25};
        \node[text=black] at (axis cs: 4,0) {75};
        \node[text=black] at (axis cs: 5,0) {43};
        \node[text=black] at (axis cs: 0,1) {38};
        \node[text=black] at (axis cs: 2,1) {33};
        \node[text=white] at (axis cs: 3,1) {84};
        \node[text=black] at (axis cs: 4,1) {54};
        \node[text=black] at (axis cs: 5,1) {54};
        \node[text=white] at (axis cs: 0,2) {94};
        \node[text=black] at (axis cs: 1,2) {33};
        \node[text=black] at (axis cs: 3,2) {2};
        \node[text=black] at (axis cs: 4,2) {75};
        \node[text=black] at (axis cs: 5,2) {39};
        \node[text=black] at (axis cs: 0,3) {25};
        \node[text=white] at (axis cs: 1,3) {84};
        \node[text=black] at (axis cs: 2,3) {2};
        \node[text=black] at (axis cs: 4,3) {46};
        \node[text=black] at (axis cs: 5,3) {29};
        \node[text=black] at (axis cs: 0,4) {75};
        \node[text=black] at (axis cs: 1,4) {54};
        \node[text=black] at (axis cs: 2,4) {75};
        \node[text=black] at (axis cs: 3,4) {46};
        \node[text=black] at (axis cs: 5,4) {55};
        \node[text=black] at (axis cs: 0,5) {43};
        \node[text=black] at (axis cs: 1,5) {54};
        \node[text=black] at (axis cs: 2,5) {39};
        \node[text=black] at (axis cs: 3,5) {29};
        \node[text=black] at (axis cs: 4,5) {55};
      \end{axis}
    \end{tikzpicture}
}
    \caption{Heatmap of similarities between data of each style in our dataset using averaged SBERT embeddings (scores are between 0 and 1, decimals omitted for ease of viewing).}
    \label{fig:heatmap}
\end{figure}
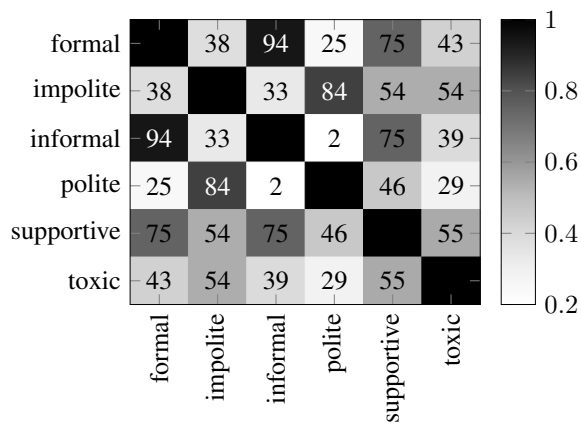

As a preliminary analysis, we measured the similarity between our datasets and their styles. We used SBERT~\cite{Reimers.2019} to calculate the sentence embedding of each instance in our dataset that belongs to the formal, informal, polite, impolite, offensive, supportive, and toxic styles. We calculated the average SBERT embedding for each style and measured the similarity between these style embeddings. The results are shown in Figure~\ref{fig:heatmap} for similarities between 0 and 1. We see some similarity between styles that come from the same datasets, e.g. informal and formal texts are very similar, as this is a parallel corpus. Aside from these similarities, we noticed similarities between styles that may generally have the same positive or negative connotation, e.g. toxic, informal, and offensive text are highly similar. This inspired the positive and negative categories for locality levels, with positive including formal, polite, and supportive text, and negative including informal, impolite, offensive, and toxic text.

\end{document}